\documentclass[english]{article}
\usepackage{spconf}

\usepackage{babel}
\usepackage[utf8]{inputenc} %
\usepackage[T1]{fontenc}

\usepackage{caption}
\usepackage{subcaption}

\usepackage{graphicx}
\usepackage{amsmath,bm}
\usepackage{amsfonts}
\usepackage{amssymb}
\usepackage{mathtools}
\usepackage{array}
\usepackage{multirow}
\usepackage{tabularx}
\usepackage{cite}
\usepackage[keeplastbox]{flushend}
\usepackage{footnote}
\usepackage{soul} % small caps / strikethrough
\usepackage{hhline} % better hline for tabluar
\usepackage{ifthen}
\usepackage{verbatim} % comment environment
\usepackage{siunitx}
\usepackage{url} % pour lien html
\usepackage[ruled, vlined, linesnumbered]{algorithm2e}
\usepackage[hidelinks]{hyperref} % provides \autoref, draft option for disabling hyperlinks
\usepackage{multicol} % multicolumn in floats
\usepackage{listings}
\usepackage[noabbrev,capitalise]{cleveref} % eq ref
\usepackage{colortbl}
\usepackage{import}
\usepackage{pgfplots}
\usepackage{adjustbox}
\usepackage{interval}

\usepackage{tikz}

\definecolor{gray}{gray}{0.4}
\definecolor{prettygreen}{rgb}{0.0,0.8,0.0}
\definecolor{darkgreen}{rgb}{0.0,0.4,0.0}
\definecolor{darkred}{rgb}{0.55, 0.0, 0.0}
\definecolor{darkblue}{rgb}{0.0, 0.0, 0.55}
\SetKwInOut{Input}{input}
\SetKwInOut{Output}{output}
\SetKwInOut{InOut}{in/out}
\SetKwInOut{Temp}{temp}
\SetKw{KwFrom}{from}

\let\envcomment\comment
\let\endenvcomment\endcomment
\renewcommand*{\comment}[2][\hspace{1.5em}]{\textsf{\smaller{#1// #2}}}

\let\oldnl\nl% Store \nl in \oldnl
\newcommand{\nonl}{\renewcommand{\nl}{\let\nl\oldnl}}% Remove line number for one line
\newcolumntype{C}{>{\centering\arraybackslash}X}
\newcolumntype{R}{>{\raggedleft\arraybackslash}X}
\newcolumntype{L}{>{\raggedright\arraybackslash}X}
\graphicspath{{img/}}
\usetikzlibrary{arrows.meta,shapes.misc}

\tikzset{cross/.style={draw,cross out, minimum size=2*(#1-\pgflinewidth), inner sep=0pt, outer sep=0pt},
cross/.default={1ex}}
\tikzset{label/.style={draw,shape=circle,inner sep=#1},
label/.default={0.1em}}
\newenvironment{map}[1]{%
    \global\expandafter\def\csname map@#1 \endcsname{}%
    \def\set##1##2{%
        \global\expandafter\def\csname map@#1@##1 \endcsname{##2}%
    }%
    \def\get##1{%
        \ifcsname map@#1@##1 \endcsname%
            \csname map@#1@##1 \endcsname%
        \else%
            \PackageError{Map}{"##1" is not defined in map "#1"}{The map "#1" exists but does not contain the key "##1". Please ensure you have typed the key correctly.}%
            ##1%
        \fi%
    }%
}{}

\newcommand{\mapget}[2]{%
    \ifcsname map@#1 \endcsname%
        \begin{map}{#1}%
            \get{#2}%
        \end{map}%
    \else%
        \PackageError{Map}{map "#1" is not defined}{Please ensure you have typed the map name correctly.}%
        #2%
    \fi%
}

\newcommand{\definemap}[1]{%
    \begin{map}{#1}%
    \end{map}%
    \expandafter\def\csname#1\endcsname{\mapget{#1}}%
}
\ifthenelse{\isundefined{\diffmode}}{
    \newenvironment{newtext}{%
        \bgroup%
    }{\egroup}
    \newenvironment{temptext}{%
        \bgroup%
    }{\egroup}
    \newenvironment{removedtext}{%
        \bgroup%
        \envcomment%
    }{% 
        \endenvcomment%
        \egroup%
    }
    
    \newcommand*{\new}[1]{{#1}}
    
    \newcommand*{\removed}[1]{{}}
    \newcommand*{\replace}[2]{{#2}}
    \newcommand*{\missing}[1]{}
}{
    \newcommand*{\new}[1]{{\color{darkgreen}#1}}
    
    \newcommand*{\removed}[1]{{\color{darkred}#1}}
    \newcommand*{\replace}[2]{\removed{#1} \new{#2}}
    \newcommand*{\missing}[1]{\raisebox{0pt}{\color{red}[{\smaller#1}]}}
}
\newcommand{\todo}[1]{\bgroup\color{red}#1\egroup}
\newcommand*{\acronym}[1]{{#1}}

\newcommand*{\unit}[1]{%
    \ifmmode%
        \ \mathrm{#1}%
    \else%
        $\mathrm{#1}$%
    \fi%
}

\newcommand*{\nbyn}[1]{\ensuremath{#1{\times}#1}}

\newcommand*\circled[2][none]{%
    \adjustbox{padding=-0.25pt}{%
        \sffamily\smaller%
        \tikz[baseline=(char.base)]{%
            \node[shape=circle,draw,inner sep=0.75pt,fill={#1}] (char) {#2};%
        }%
    }%
}
\newcommand{\bglabel}[1]{\circled[white]{#1}}
\newcommand{\fglabel}[1]{\circled[lightgray]{#1}}

\newcommand*{\includepgf}[3][]{%
  \bgroup%
  \newcommand{\setmainfont}[1]{}%
  \let\pgfimage\includegraphics% % \pgfimage does not take into account the subimport
  \adjustbox{#1}{\subimport*{#2/}{#3.pgf}}%
  \egroup%
}
\definemap{mset}
\begin{map}{mset}
    \set{N}{\ensuremath{\mathbb{N}}}
    \set{Z}{\ensuremath{\mathbb{Z}}}
    \set{D}{\ensuremath{\mathbb{D}}}
    \set{Q}{\ensuremath{\mathbb{Q}}}
    \set{R}{\ensuremath{\mathbb{R}}}
    \set{C}{\ensuremath{\mathbb{C}}}
    
    \set{S}{\ensuremath{\mathbb{S}}}
    \set{S+}{\ensuremath{\get{S}_+}}
    \set{S++}{\ensuremath{\get{S}_{++}}}
    \set{L}{\ensuremath{\mathbb{L}}}
    \set{Lu}{\ensuremath{\get{L}_u}}
\end{map}

\definemap{simd}
\begin{map}{simd}
    \set{}{\acronym{SIMD}}
    \set{sse}{\acronym{SSE}}
    \set{sse2}{\acronym{SSE2}}
    \set{sse3}{\acronym{SSE3}}
    \set{ssse3}{\acronym{SSSE3}}
    \set{sse4}{\acronym{SSE4}}
    \set{sse4.1}{\acronym{SSE4.1}}
    \set{sse4.2}{\acronym{SSE4.2}}
    \set{avx}{\acronym{AVX}}
    \set{avx2}{\acronym{AVX2}}
    \set{avx512}{\acronym{AVX512}}
    \set{neon}{Neon}
    \set{sve}{\acronym{SVE}}
    \set{altivec}{Altivec}
    \set{vmx}{\acronym{VMX}}
    \set{vsx}{\acronym{VSX}}
\end{map}

\title{a new run-based Connected Component Labeling for efficiently analyzing and processing holes}
\author{Florian Lemaitre and Lionel Lacassagne}
\date{January 2020}

\name{Florian Lemaitre and Lionel Lacassagne}
\address{Sorbonne University, CNRS, LIP6, France\\fisrtname.name@lip6.fr}

\begin{document}
\ninept % Si on veut plus de place, on peut passer en 9pt

\maketitle

\begin{abstract}
This article introduces a new connected component labeling and analysis algorithm for foreground and background labeling that computes the adjacency tree.
The computation of features (bounding boxes, first statistical moments, Euler number) is done on-the-fly.
The transitive closure enables an efficient hole processing that can be filled while their features are merged with the surrounding connected component without the need to rescan the image.
A comparison with existing algorithms shows that this new algorithm can do all these computations faster than algorithms processing black and white components.
\end{abstract}

% =========================================
\section{Introduction \& State of the Art}
% =========================================

Connected Component Labeling (CCL) is a fundamental algorithm in computer vision.
It consists in assigning a unique number to each connected component of a binary image.
Since Rosenfeld \cite{Rosenfeld1966}, many algorithms have been developed to accelerate its execution time on CPU \cite{Cabaret2014_CCL_SIPS, He2017_review_PR, Grana2019_spaghetti}, SIMD CPU \cite{Hennequin2019_SIMD_CCL_WPMVP}, GPU \cite{Playne2018} or FPGA \cite{Klaiber2016_JPRTIP}.
% AJOUTER LEMAITRE2020

%In the same time Connected Component Analysis (CCA) that consists in computing CC features -- like bounding-box to extract characters for OCR, or the first raw moments ($S$, $S_x$, $S_y$) for motion detection and tracking -- has also rise \cite{Bailey2007} \cite{Lacassagne2009_LSL_ICIP}\cite{Lacassagne2011_LSL_JRTIP}\cite{Tang2016_run_CCA}\cite{He2019_Features}. Parallelized algorithms have been also designed \cite{Cabaret2018_parallel_LSL_JRTIP} \cite{Hennequin2018_GPU_CCL_DASIP} \cite{Bailey2019_zig_zag}.\\

% on gagne une ligne en supprimant les crochets intermediaires
In the same time Connected Component Analysis (CCA) that consists in computing Connected Component (CC) features -- like bounding-box to extract characters for OCR, or the first raw moments ($S$, $S_x$, $S_y$) for motion detection and tracking -- has also risen \cite{Bailey2007, Lacassagne2009_LSL_ICIP, Lacassagne2011_LSL_JRTIP, Tang2016_run_CCA, He2019_Features}.
Parallelized algorithms have been also designed \cite{Cabaret2018_parallel_LSL_JRTIP, Hennequin2018_GPU_CCL_DASIP, Bailey2019_zig_zag}.
The initial Union-Find algorithm \cite{Tarjan1975} has been also analysed \cite{Tarjan1984} and improved \cite{Galil1991} with decision tree \cite{Wu2009} and various path compression/modification algorithms \cite{Manne2009, Patwary2010}.

%enlever Tarjan1984, Leeuwen1977, et Manne2009 si cela deborde

Some other features -- useful for pattern classification/recognition -- are computed by another set of algorithms: the Euler number with Bit-Quads \cite{Gray1971_bitquad}, the adjacency (also known as homotopy or inclusion) tree \cite{Rosenfeld1979_adjacency_tree} and more recently, foreground (FG) and background (BG) labeling (also known as B\&W labeling) \cite{He2012_HCS_hole} and hole filling \cite{He2012_HCS_hole} with also improvements in the last decade: \cite{He2017_bitquad_unroll_euler, DiazDelRio2019_adjacency_tree}. 

\Cref{sec:lsl} provides a short description of LSL algorithm that is used for our new algorithm, \cref{sec:lsl_hole} introduces our new algorithm, \cref{sec:benchmark_performance_analysis} presents a benchmark of existing algorithms and their analysis and \cref{sec:conclusion} is the conclusion.

\section{Classical LSL} \label{sec:lsl}
% ===================================
%The basis of our new algorithm is standard \acronym{CCL} algorithms.
%We chose to start from LSL\cite{Lacassagne2011_LSL_JRTIP} as it is able to compute features very quickly due to its segment nature.
We chose to base our new algorithm on LSL \cite{Lacassagne2011_LSL_JRTIP} because it is run-based (segment processing) and thus is able to compute features very quickly compared to pixel-based algorithms.
% ------------
\begin{figure}
% ------------
\centering
    \includegraphics[scale=0.70]{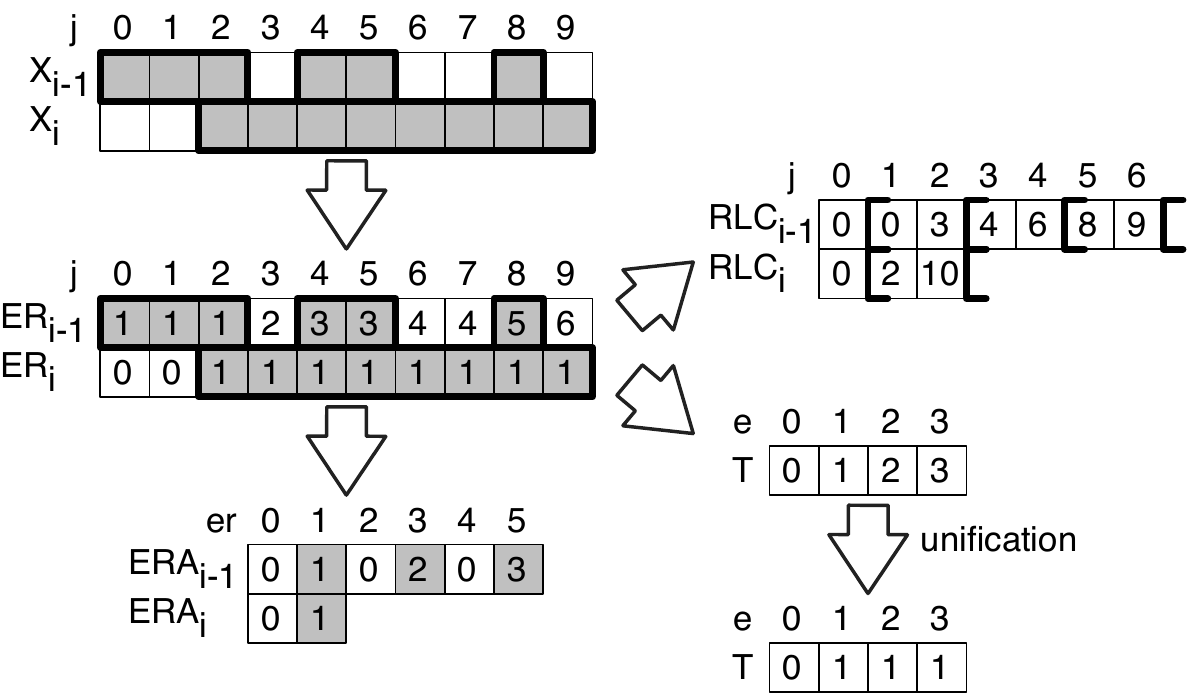}
    \caption{LSL tables for classic foreground labeling: the \emph{even} relative labels corresponding to background components are equivalent to zero in $ERA$ tables (they will be equivalent to other labels for the proposed foreground and background labeling).}
    \label{fig:LSL_tables}
\end{figure}

The specificity of LSL is to combined segments with line-relative labeling to speed segment adjacency detection up.
Like all direct CCA algorithms, it is split into two steps and performs a single image scan.

The first step assigns a temporary/provisional label to each connected component and computes its associated features.
Equivalences between labels are built when needed.
It is composed of a segment detection (\Cref{algo:rle:scalar}) and a segment Unification (\Cref{algo:lsl:unification} modified for B\&W labeling).
From two consecutive input lines $X_{i-1}$, $X_i$, two relative labelings ($\mathit{ER}_i$, $\mathit{ER}_{i-1}$) are produced where FG runs (or segments) have odd numbers and BG runs have even numbers.
The associated semi-open intervals are stored in tables $\mathit{RLC}_{i-1}$, and $\mathit{RLC}_i$.
The table $\mathit{ERA}_{i}$ holds the translation between \emph{Relative} and \emph{Absolute} labels: $\mathit{ea} = \mathit{ERA}_i[er]$.

To find out which labels of the previous line are connected to the current segment (Unification), one has to read the value of relative labels from table $\mathit{ER}_{i-1}$ at the positions given by $\mathit{RLC}_i$, and translates them into absolute labels to update the equivalence table~{$T$}.
\replace{During this step, features -- that are stored in $F$ table -- are also computed.}{%
Features are also computed during this step and stored in table~{$F$}.
}% and updated according to the equivalence building.\todo{the merge of features is done during TC}% The Unification can also be done by a finite state machine that may be faster on some architectures. 

The second step solves the equivalence table by computing the transitive closure (TC) of the graph associated to the label equivalence and merge the features together.
Unlike for CCL algorithms, the third step that performs a second scan to replace temporary labels of each connected component with their final labels (\Cref{algo:relabel}) is usually avoided (but can be done for visual verification).

In order to differentiate \emph{temporary} root and \emph{final} root, we define $a$ ($a \leftarrow \operatorname{Find}(e_a)$) as the \emph{temporary} root of the absolute label $e_a$ used in \Cref{algo:lsl:unification}, while $r$ used in \Cref{algo:closure} and \Cref{algo:relabel} is the \emph{final} root of the connected component.

% \todo{YA ENCORE DE LA PLACE}
% \textcolor{darkgreen}{bla bla bla bla bla bla bla bla bla bla bla bla bla bla bla bla bla bla bla bla bla bla bla bla bla bla bla bla bla bla bla bla bla bla bla bla bla bla bla bla bla bla bla bla}

%More details of the original LSL are available in \cite{Lacassagne2011_LSL_JRTIP}.\\ 

% ---------------
\begin{algorithm}
% ---------------
\DontPrintSemicolon
\begin{footnotesize}
%\KwIn{$X_i$ a binary line of width $w$}
%\KwResult{$ER_i$, $RLC_i$ and $ner$}
\comment[]{prolog} \;
$\mathit{RLC}_{i}[0] \leftarrow 0$ \;
$\mathit{er} \leftarrow 0$ \;
$x_{1} \leftarrow 0$ \comment{previous value of $X$ ($x_{1}=X_i[j-1]$)}\;
\For{$j=0$ \KwTo $w-1$}{
    $x_0 \leftarrow X_i[j]$ \comment{current value} \;
    $\mathit{RLC}_{i}[\mathit{er}+1] \leftarrow j$ \comment{will later be overwritten if $f = 0$} \;
    $f \leftarrow x_{0} \oplus x_{1}$ \comment{edge detection} \;
    $ \mathit{er} \leftarrow \mathit{er} + f$ \;
    $ \mathit{ER}_i[j] \leftarrow \mathit{er} $ \;
    $x_{1} \leftarrow x_0$ \comment{register rotation}
}
\comment[]{epilog} \;
$\mathit{RLC}_i[\mathit{er}+1] \leftarrow w$ \;
$ \mathit{ner}_i \leftarrow er + 1$ \;
\Return{$\mathit{ner}_i$} \;
\end{footnotesize}
\caption{segment detection (LSL step 1a)}
%\label{algo:LSL_segment_detection_STD}
\label{algo:rle:scalar}
\end{algorithm}

% % ------------------
% \begin{algorithm}[t]
% % ------------------
% \DontPrintSemicolon
% \begin{footnotesize}
% %\KwIn{$X_i$ a binary line of width $w$}
% %\KwResult{$ER_i$, $RLC_i$ and $ner$}
% \comment[]{prolog} \;
% $er \leftarrow 0$ \;
% $x_1 \leftarrow 0$ \comment{previous value of $X$ ($x_1=X_i[j-1]$)}\;
% \For{$j=0$ \KwTo $w-1$}{
%     $x_0 \leftarrow X_i[j]$ \comment{current value} \;
%     $f \leftarrow x_{0} \oplus x_{1}$ \comment{edge detection} \;
%     \If{$f \neq 0$} { 
%         $RLC_{i}[er] \leftarrow j$ \;
%         $er \leftarrow er + 1$ \;
%     }
%     $ER_i[j] \leftarrow er$ \;
%     $x_1 \leftarrow x_0$ \comment{register rotation}
% }
% \comment[]{epilog} \;
% $x_0 \leftarrow 0$ \comment{next value of} $X$ ($x_1=X_i[w]$)\;
% $f \leftarrow x_0 \oplus x_1$ \;
% $RLC_i[er] \leftarrow w$ \;
% $er \leftarrow er + f$ \;
% $ner \leftarrow er$ \;
% \Return{$ner$} \;
% \end{footnotesize}
% \caption{Segment detection (step 1a)}
% %\label{algo:LSL_segment_detection_RLC}
% \label{algo:rle:scalar}
% \end{algorithm}

% ====================================================
\section{LSL and Hole processing} \label{sec:lsl_hole}
% ====================================================
Our goal is to propose a new \emph{all-in-one} algorithm that finds holes and processes them efficiently by computing the adjacency tree between foreground and background connected components.
%
%The goal of our new algorithm is to find holes and process them efficiently by computing the adjacency tree between foreground and background connected components.
%
Then, filling holes only consists in modifying the adjacency tree and the equivalence table, without modifying pixels or labels in the image (the same approach can be used to apply arbitrary connected operators\cite{Salembier2009_connected_operators}).
\replace{Our algorithm must also be able to}{We also want our algorithm to be able to} compute statistical features on-the-fly, to relabel the whole image and to compute adjacency features like the Euler number -- or just a subset.
%all of them.
% Moreover the statistical features can be computed on the fly. We can also choose  to label the whole image or just to compute some features of the adjacency tree like Euler number.

%We can also choose to compute \acronym{CC} features on-the-fly, to label the whole image or just to compute some features of the adjacency tree like Euler number.

%cette aprocah peut aussi etre utilisée pour plein d'autres truc

%The goal of our new algorithm is to find the holes and the relations between holes and foreground \acronym{CC}s in order to virtually change pixels of the holes from background to foreground without doing any extra scan of the image.
% This is done by computing both foreground and background \acronym{CC}s as well as the adjacency tree of these \acronym{CC}s whose root is the exterior of the image (implicitly background).
% Arbitrary connected operators\cite{Salembier2009_connected_operators} can then be applied by modifying this tree.
% Hole filling is a useful example of connected operators \todo{and the only one considered here ?}.
%Filling holes is then just a \emph{manipulation} of this tree.

In the following, a ``component'' means a Connected Component, either foreground or background.

% -------------------------------------------------------
%\subsection{Holes, surrounding and adjacency tree}
\subsection{Holes and adjacency tree}
% -------------------------------------------------------

% % ------------
% \begin{figure}
% % ------------
%     \centering
%     %\includegraphics[scale=0.70]{figures/closing_pattern.pdf}
%     %\includegraphics[scale=0.35]{figures/patterns.pdf}
%     %\includegraphics[scale=0.35]{figures/patterns2.pdf}
%     \includegraphics[scale=0.35]{figures/patterns2.pdf}
%   %\caption{Closing patterns, for 8C (if $a$ or $d \equiv c$), 4C (if $a \equiv b$) and for segment (if $a \equiv b$)}
%   %\caption{Three patterns of closing pixels (if $a$ or $d \equiv c$) and one pattern of surrounding pixels ($a \equiv b \equiv c \equiv d$) for 8-foreground and 4-background. \todo{AJOUTER LES SCHEMAS DE FLORIAN + dire initial adjacency}}
%   \caption{Four \emph{closing pixels} patterns (if $a$ or $d \equiv c$ in 8C and if $b \equiv d$ in 4C)  (top line), two \emph{initial adjacency} patterns  (bottom line). With for each line, 8-foreground and 4-background on left, 4-foreground and 8-background on right}
%   \label{fig:surrounding_closing_pixels}
% \end{figure}

A component $C_1$ is surrounded by another component $C_2$ -- written $C_1 \sqsubset C_2$ -- if and only if all paths from $C_1$ to the exterior of the image contain at least one pixel from $C_2$.
A hole is a background component that is surrounded by a foreground component.
%\todo{PAS CLAIR A hole in a foreground \acronym{CC} $W$ is a background \acronym{CC} $B$ that is surrounded by $W$}.

The adjacency tree is encoded in a new table $I$.
For a label $e_1$ associated to a component $C_1$, $e_2 = I[e_1]$ is one of the temporary labels of the unique component $C_2$ that is both adjacent to $C_1$ and surrounding $C_1$ ($C_1 \sqsubset C_2$).
%The label $e_2$ is not necessarily a root label.
$I[e_1] = -1$ if $e_1 = 0$, or $e_1$ is not a root label ($T[e_1] \neq e_1$).
In short, the table $I$ represents the adjacency tree whose edges are directed according to the surrounding relation.

We considered two methods to build the adjacency tree and the surrounding relation: detecting closing pixels\cite{He2015_closing_pixel}, or looking at the adjacency at exterior pixels\cite{DiazDelRio2019_adjacency_tree}.\medskip

\begin{figure}
% ------------
    \newcommand{\figscale}{0.35}%
    \setlength{\tabcolsep}{0.5em}%
    \sffamily%
    \subcaptionbox{Closing pixel patterns\label{fig:surrounding_patterns:closing_pixel}}{%
        \begin{tabular}{@{}lll@{}}
            FG: &
            \adjustbox{valign=m,stack=l}{$a \equiv c\,$ or \\ $d \equiv c$} &
            \adjincludegraphics[scale=\figscale,valign=m]{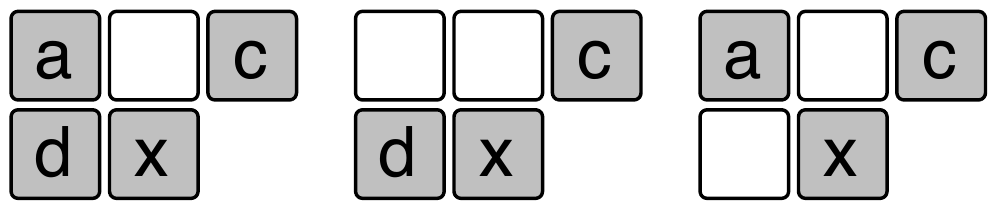} \\\\
            BG: &
            $d \equiv b$ &
            \adjincludegraphics[scale=\figscale,valign=m]{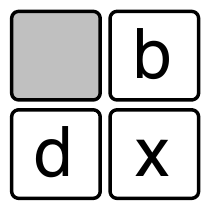}
        \end{tabular}%
    }\hfill%
    \subcaptionbox{New label patterns\label{fig:surrounding_patterns:new_label}}{%
        \hspace{2em}%
        \begin{tabular}{@{}ll@{}}
            FG: &
            \adjincludegraphics[scale=\figscale,valign=m]{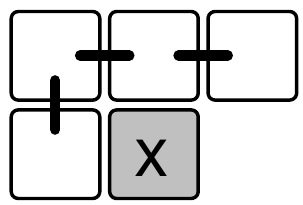} \\\\
            BG: &
            \adjincludegraphics[scale=\figscale,valign=m]{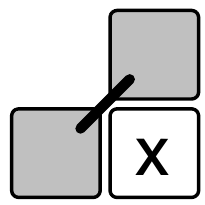}
        \end{tabular}%
    }%
    \caption{Patterns demonstrating how surroundings are detected when the current pixel \textsf{x} is processed. Foreground is 8-adjacent and background is 4-adjacent.}
    \label{fig:surrounding_patterns}
\end{figure}

%We can detect that $A$ and $B$ are adjacent and $A \sqsubset B$ by detecting closing pixels \cite{He2015_closing_pixel}.
A closing pixel is a pixel neighboring both ends of a path (using 4-adjacency for BG and 8-adjacency for FG).
\Cref{fig:surrounding_patterns:closing_pixel} shows the patterns of a closing pattern while it is being processed.
The surrounding $C_1 \sqsubset C_2$ where $C_1$ and $C_2$ are adjacent can be detected during the Unification when $C_2$ is unified with itself.
If $C_2$ is FG (8-adjacency), the pixel above the closing pixel is in $C_1$.
If $C_2$ is BG (4-adjacency), the upper-left pixel is in $C_1$.\medskip

% ------------

%\vspace*{-2em}%
The other method relies on the adjacency at exterior pixels.
In our case, we consider the top most pixels of a component because those require the creation of a new label which is easy to detect.
Every time a new label is created, the label directly above the current pixel is recorded in $I$ as its initial adjacency and speculative surrounding.
It is actually simpler to look for the label on the left that is necessarily from the same component as above (\Cref{fig:surrounding_patterns:new_label}).
When two labels $a < b$ are unified, the initial adjacency $I[b]$ is discarded in favor of $I[a]$.
The order on labels implies that top pixels of $a$ are higher than top pixels of $b$ -- or at least at the same height.
It means that the higher initial adjacency and speculative surrounding is kept while the other is discarded.
Once a component has been fully scanned, the only initial adjacency kept for this component is the one from the root label which is, by construction, the label of top most pixels, and thus on the exterior of the component.
The remaining initial adjacency and speculative surrounding is thus necessarily a true surrounding.\medskip
% Another way to compute $I$ is to rely on ``\emph{initial adjacent pixels}''.
% Every time a new label is created, the label on its left is recorded in $I$ as its initial adjacency.
% To be noted that the label on the left and the one above are equivalent. %\todo{(\Cref{fig:new_label_adjacency})}.
% When two labels $a < b$ are unified, the initial adjacency $I[b]$ is discarded in favor of $I[a]$.
% The order on labels implies that top pixels of $a$ are higher than top pixels of $b$ -- or at least at the same height.
% It means that the higher initial adjacency is kept while the other is discarded.
% Once the image is fully scanned, only remains the initial surroundings of the top most pixels of each component.
% As they are on top, those pixels are necessarily on the exterior of their component, and pixels above are their actual surrounding -- which is what is encoded within $I$.\medskip

%It means that $I[r]$ encodes by construction not only the adjacency of a root label $r$, but also the surrounding component of $r$.

We chose to use the initial adjacency method as it saves one extra branch and one extra $\operatorname{Find}$ within the Unification compared to the closing pixel method.
Moreover, the update of $I$ when an adjacency is discarded is actually not necessary as $I$ is accessed only for root labels whose initial adjacencies are kept by construction.
While the adjacency is a local property, the surrounding is not and thus is defined and correct only when the component has been fully scanned.
Consequently, initial adjacency builds a speculative $I$ that is correct only at the end of the image scan and that cannot be worked on beforehand.

%In order to figure out what part of the algorithm is important (that makes one algorithm to be faster than other ones), we modified the LSL: there are still the STD (standard) and RLE (RLE compression that uses $LEA$ tables) versions, but the equivalence management can be either the classic Union-Find (named Rosenfeld in the following) or the Suzuki method. \\

%The first step is a two-fold step: there is first a segment detection (\Cref{algo:LSL_segment_detection_RLC}) algorithm (based on RLE with semi-open interval) that also generates line-relative labels.  Then a second algorithm build the equivalence between absolutes labels of  $i$ and $i-1$ (\Cref{algo:LSL_unification}). An alternate algorithm is the finite state machine of \cref{fig:LSL_automate}. The unification is accelerated thanks to RLE: only one test per segment instead of one per pixel. The second step is the same than pixel-based algorithm (\Cref{algo:solve}). The third step is also RLE-accelerated: it is a simple RLE decompression where each label is replaced by its root (\Cref{algo:LSL_update}).

% ------------------
\subsection{New Black and White LSL}
% ------------------

% ---------------
\begin{algorithm}
% ---------------
\DontPrintSemicolon
\begin{footnotesize}
%\SetKw{\KwReturn}{return}
%\SetKwInOut{Require}{Require}
%\KwIn{$ER$, $RLC$, $EQ$, $ERA$, $ner$}
%\KwResult{$ne$ the current number of absolute labels, update of $T$ and $ERAi$}
\SetKw{KwStep}{step}
\SetKw{KwIs}{is}
$\mathit{er}_s \leftarrow 0$ \comment{starting segment}\label{algo:lsl:unification:prolog-begin}\;
\If{$\mathit{RLC}_i[1] = 0$}{
    $\mathit{ERA}_i[0] \leftarrow 0$ \;
    $\mathit{er}_s \leftarrow 1$ \;
}
\comment[]{White-only would have $\mathit{er}_s = 1$ and step = 2}\label{algo:lsl:unification:prolog-end} \;
\For{$\mathit{er}=\mathit{er}_s$ \KwTo $\mathit{ner}_i-1$ \KwStep $1$}{
    \comment[]{semi open interval segment extraction $\interval[open right]{j_0}{j_1}$}\;
    $j_0 \leftarrow \mathit{RLC}_i[\mathit{er}]$\; 
    $j_1 \leftarrow \mathit{RLC}_i[\mathit{er}+1]$\;
    $p \leftarrow \mathit{er} \mathbin{\operatorname{mod}} 2$ \comment{parity of current segment} \;
    $ c_8 \leftarrow p$\label{algo:lsl:unification:c8} \comment{if current segment is 8-C, $c_8=1$} \;
    $ f \leftarrow \operatorname{ComputeFeatures}(i, j_0, j_1)$ \;
    %$j_0 \leftarrow RLC_i[er-1], \quad j_1 \leftarrow RLC_i[er]-1$\; 
    \comment[]{fix extension in case of 8-connected component}\label{algo:lsl:unification:fix_j} \;
    $ j_0 \leftarrow \operatorname{max}(j_0 - c_8, 0)$ \;
    $ j_1 \leftarrow \operatorname{min}(j_1 + c_8, w)$ \;
    $\mathit{er}_0 \leftarrow \mathit{ER}_{i-1}[j_0] $\; 
    $\mathit{er}_1 \leftarrow \mathit{ER}_{i-1}[j_1-1]$ \comment{right compensation} \;
    %$er_0 \leftarrow ER_{i-1}[j_0], \quad e_{r1} \leftarrow ER_{i-1}[j_1]$ \;
    \comment[]{fix label parity: BG segments are even, FG segments are odd}\label{algo:lsl:unification:fix_er} \;
    $\mathit{er}_0 \leftarrow \mathit{er}_0 + ((er_0 \mathbin{\operatorname{mod}} 2) \oplus p)$ \;
    $\mathit{er}_1 \leftarrow \mathit{er}_1 - ((er_1 \mathbin{\operatorname{mod}} 2) \oplus p)$ \;

    \eIf{$\mathit{er}_1 \ge \mathit{er}_0$}{
        $e_a \leftarrow \mathit{ERA}_{i-1}[\mathit{er}_0]$ \;
        $a \leftarrow \operatorname{Find}(e_a)$ \;
        \For{$\mathit{er}_k=\mathit{er}_0+2$ \KwTo $\mathit{er}_1$ \KwStep $2$}{
            $\mathit{ea}_k \leftarrow \mathit{ERA}_{i-1}[\mathit{er}_k]$ \;
            $a_k \leftarrow \operatorname{Find}(\mathit{ea}_k)$ \;
            \If(\comment{Union: min propagation}){$a < a_k$}{
                $T[a_k] \leftarrow a$ \;
            }
            \If(\comment{Union: min extraction}){$a > a_k$}{
                $T[a] \leftarrow a_k$ \;
                $a     \leftarrow a_k$ \;
            }
        }
        $\mathit{ERA}_i[er] \leftarrow a$ \comment{the global min} \;
        $F[a] \leftarrow F[a] \cup f$ \comment{update features} \;
    }(\comment{new label}){
        $\mathit{ERA}_i[\mathit{er}] \leftarrow \mathit{ne}$ \;
        $F[\mathit{ne}] \leftarrow f$ \;
        $I[\mathit{ne}] \leftarrow \mathit{ERA}_i[\mathit{er}-1]$ \comment{Initial adjacency} \label{algo:lsl:unification:initial-adjacency}\;
        $\mathit{ne} \leftarrow \mathit{ne} + 1$  \;
        %$ne \leftarrow ne + 1, \quad ERAi[er/2] \leftarrow ne \quad$ \textsf{[new label]}\;
    }
}
\comment[]{first and last BG segments shall be connected to $0$} \label{algo:lsl:unification:epilog-begin}\;
$a \leftarrow \operatorname{Find}(\mathit{ERA}_i[0])$ \;
$T[a] \leftarrow 0$ \;
\If(\comment{last segment is BG}){$\mathit{ner}_i$ \KwIs odd}{
    $a \leftarrow \operatorname{Find}(\mathit{ERA}_i[\mathit{ner}_i-1])$ \;
    $T[a] \leftarrow 0$ \label{algo:lsl:unification:epilog-end} \;
}
\end{footnotesize}
\caption{B\&W Unification (step 1b)}
\label{algo:lsl:unification}
\end{algorithm}

% ---------------
\begin{algorithm}
% ---------------
\DontPrintSemicolon
\SetKw{KwAnd}{and}
\begin{footnotesize}
\For{$e = 0$ \KwTo $\mathit{ne} - 1$}{
    $ a \leftarrow T[e]$ \comment{ancestor} \;
    \If(\comment{If label is root}\label{algo:closure:hole-filling-begin}){Hole filling \KwAnd $e = a$}{
        $i \leftarrow I[e] $ \comment{label of the surrounding component} \;
        \lIf{$T[i] > 0$}{$a \leftarrow T[e] \leftarrow i$\label{algo:closure:hole-filling-end}}
    }
    \eIf{a < e}{
        
        $r \leftarrow T[a]$ \;    
        $T[e] \leftarrow r$ \comment{Transitive Closure} \;
        $F[r] \leftarrow F[r] \cup F[e]$ \comment{Feature merge} \;
    }(\comment{$e$ is a root}){
        $I[e] \leftarrow T[I[e]]$ \comment{point adjacency to root}\label{algo:closure:adjacency}\;
    }
}
\end{footnotesize}
\caption{B\&W Transitive closure (step 2)}
\label{algo:closure}
\end{algorithm}

% ---------------
\begin{algorithm}
% ---------------
\DontPrintSemicolon
\begin{footnotesize}
\For{$i=0$ \KwTo $h-1$}{
    $ j_0 \leftarrow \mathit{RLC}_{i}[0] $ \comment{$j_0$ is $0$}\;
    \comment[]{White-only would have have step = 2 and unconditionally store the BG label for even $\mathit{er}$}\;
    \For{$\mathit{er} = 0$ \KwTo $\mathit{ner}_i - 1$ \KwSty{step} $1$}{
        $e \leftarrow \mathit{ERA}_{i}[er]$ \comment{provisional label}\;
        $r \leftarrow T[e]$ \comment{final label} \;
        $ j_1 \leftarrow \mathit{RLC}_{i}[\mathit{er}+1]$ \;
        $ E_{i}\interval[open right]{j_0}{j_1} \leftarrow r$ \;
        $ j_0 \leftarrow j_1 $ \;
    }
}
\end{footnotesize}
\caption{B\&W  Relabeling (step 3)}
\label{algo:relabel}
\end{algorithm}

%deja dit avant
%The basis of our new algorithm is standard \acronym{CCL} algorithms.
%We chose to start from LSL\cite{Lacassagne2011_LSL_JRTIP} as it is able to compute features very quickly due to its segment nature.

Our new algorithm -- LSL BW -- extends the unification and the transitive closure steps to support B\&W labels and the adjacency tree (\Cref{algo:lsl:unification} and \cref{algo:closure}).
B\&W labels require to process both odd and even segments instead of just odd ones (in \cref{algo:lsl:unification} and \ref{algo:relabel}).
Consequently, $ERA_i$ table does not necessarily have zeros at even indices anymore.
The first encoded segment is always a BG one, but might have 0 length if the first pixel is FG.
Thus, the unification needs a prolog (\Cref{algo:lsl:unification}, lines \ref{algo:lsl:unification:prolog-begin} to \ref{algo:lsl:unification:prolog-end}) to skip the first segment if empty, and an epilog (\Cref{algo:lsl:unification}, lines \ref{algo:lsl:unification:epilog-begin} to \ref{algo:lsl:unification:epilog-end}) to attach the first segment and the last segment if it is BG to the exterior.
The transitive closure need no modification to handle Black\&White labels.

In addition, the correction of coordinate and indices are now dependant on the parity of the segment being processed (\Cref{algo:lsl:unification}, lines \ref{algo:lsl:unification:fix_j} and \ref{algo:lsl:unification:fix_er}).
Note that \cref{algo:lsl:unification} implements the algorithm using 8\babelhyphen{nobreak}adjacency for FG and 4-adjacency for BG.
The line \ref{algo:lsl:unification:c8} should be modified to $c_8 \leftarrow p \oplus 1$ if one wants the complementary adjacency (4-FG, 8-BG).

The construction of the adjacency table requires only a few modifications to the algorithms to set the initial adjacency (\Cref{algo:lsl:unification} line \ref{algo:lsl:unification:initial-adjacency}) and to convert temporary labels into final labels within the $I$ table (\Cref{algo:closure} line \ref{algo:closure:adjacency}).
Hole filling is done during the transitive closure (\Cref{algo:closure}, lines \ref{algo:closure:hole-filling-begin} to \ref{algo:closure:hole-filling-end}).

%The extracted features are the bounding boxes and the first statistical moments $S$, $S_x$, $S_y$. They are stored in a new table $F$.
Like for classical LSL, the computed features for each FG and BG components are the bounding-box and the first statistical moments $S$, $S_x$, $S_y$.% They are stored in a new table $F$ % deja dit

%Note that the new B\&W Unification supports both 8-foreground with 4-background and 4-foreground with 8-background.

%\todo{Mieux decrire LSL Unification = ajouter des details, comme fg, la connexite 8C et 4C}

%\textcolor{darkgreen}{bla bla bla bla bla bla bla bla bla bla bla bla bla bla bla bla bla bla bla bla bla bla bla bla bla bla bla bla bla bla bla bla bla bla bla bla bla bla bla bla bla bla bla bla bla bla bla bla bla bla bla bla bla bla bla bla bla bla bla bla bla bla}

% ------------------
\subsection{Example}
% ------------------

% % ------------
% \begin{figure}[t]
% % ------------
% \centering
%     %\includegraphics[scale=0.70]{figures/hole4.pdf}
%     \includegraphics[scale=0.70]{figures/hole4_short.pdf}
%     \caption{Example of B\&W labeling focusing on equivalences building and adjacency setting.}
%     \label{fig:lsl_bw}
% \end{figure}
% ------------
\begin{figure}[t]
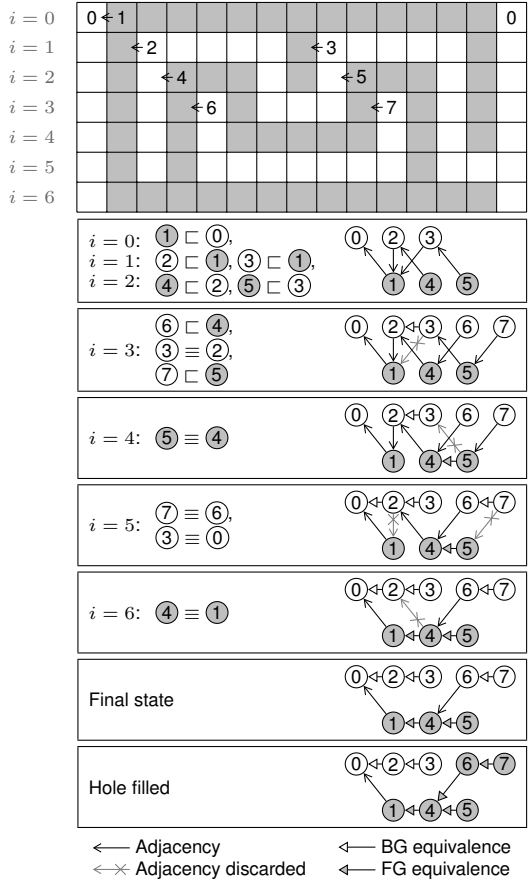

% ------------
\centering
    \sffamily\centering
    \definecolor{bg}{gray}{1}
    \definecolor{fg}{gray}{0.75}
    \renewcommand*\circled[2][none]{%
        %\smaller%
        \adjustbox{padding=-0.033em}{%
            \tikz[baseline=(char.base)]{%
                \node[label,fill={#1}] (char) {#2};%
            }%
        }%
    }
    \newcommand{\SurroundedBy}[2]{\draw[-{Straight Barb[angle=60:2pt 3]}] (#1) -> (#2);}%
    \newcommand{\NotSurroundedBy}[2]{%
        \begin{scope}[color=black!50!white]%
            \SurroundedBy{#1}{#2}%
            \path (#1) -- (#2) node[near start,sloped,draw,cross out,inner sep=0.5ex] {};
        \end{scope}%
    }%
    \newcommand{\Equiv}[3][none]{%
        \bgroup%
            %\definecolor{fg}{gray}{0.}%
            \draw[-{Triangle[fill=#1]}] (#2) -> (#3);%
        \egroup%
    }%
    
    \scriptsize
    
    \newsavebox{\tikzfigure}
    \begin{lrbox}{\tikzfigure}
    \begin{tabular}{|@{ }l@{ }l@{\hspace{1em}}l@{ }|}
        \hline
        \adjustbox{stack=l,valign=m,margin=2pt}{$i = 0$: \\ $i = 1$: \\ $i = 2$:} &
        \adjustbox{stack=l,valign=m,margin=2pt}{%
            $\circled[fg]1 \sqsubset \circled[bg]0$, \\%
            $\circled[bg]2 \sqsubset \circled[fg]1$, %
            $\circled[bg]3 \sqsubset \circled[fg]1$, \\%
            $\circled[fg]4 \sqsubset \circled[bg]2$, %
            $\circled[fg]5 \sqsubset \circled[bg]3$%
        } &
        \adjustbox{valign=m,margin=2pt}{\tikz[x=1em,y=1em]{\input{tikz/graph012.tex}}}\\\hline\hline
        
        \adjustbox{valign=m,margin=2pt}{$i = 3$:} &
        \adjustbox{stack=l,valign=m,margin=2pt}{%
            $\circled[bg]6 \sqsubset \circled[fg]4$,\\%
            $\circled[bg]3 \equiv \circled[bg]2$,\\%
            $\circled[bg]7 \sqsubset \circled[fg]5$%
        } &
        \adjustbox{valign=m,margin=2pt}{\tikz[x=1em,y=1em]{\input{tikz/graph3.tex}}}\\\hline\hline
        
        \adjustbox{valign=m,margin=2pt}{$i = 4$:} &
        \adjustbox{stack=l,valign=m,margin=2pt}{%
            $\circled[fg]5 \equiv \circled[fg]4$%
        } &
        \adjustbox{valign=m,margin=2pt}{\tikz[x=1em,y=1em]{\input{tikz/graph4.tex}}}\\\hline\hline
        
        \adjustbox{valign=m,margin=2pt}{$i = 5$:} &
        \adjustbox{stack=l,valign=m,margin=2pt}{%
            $\circled[bg]7 \equiv \circled[bg]6$,\\%
            $\circled[bg]3 \equiv \circled[bg]0$%
        } &
        \adjustbox{valign=m,margin=2pt}{\tikz[x=1em,y=1em]{\input{tikz/graph5.tex}}}\\\hline\hline
        
        \adjustbox{valign=m,margin=2pt}{$i = 6$:} &
        \adjustbox{stack=l,valign=m,margin=2pt}{%
            $\circled[fg]4 \equiv \circled[fg]1$%
        } &
        \adjustbox{valign=m,margin=2pt}{\tikz[x=1em,y=1em]{\input{tikz/graph6.tex}}}\\\hline\hline

        \multicolumn{2}{|@{ }l}{\adjustbox{valign=m,margin=2pt}{Final state}} &
        \adjustbox{valign=m,margin=2pt}{\tikz[x=1em,y=1em]{\input{tikz/graph-end.tex}}}\\\hline\hline
        
        \multicolumn{2}{|@{ }l}{\adjustbox{valign=m,margin=2pt}{Hole filled}} &
        \adjustbox{valign=m,margin=2pt}{\tikz[x=1em,y=1em]{\input{tikz/graph-filled.tex}}}\\\hline
    \end{tabular}%
    \end{lrbox}
    \newlength{\figurewidth}
    \setlength{\figurewidth}{\dimexpr\wd\tikzfigure / 15\relax}
    
    \begin{tabular}{r}
    \tikz[x=\figurewidth,y=\figurewidth]{\begin{scope}[yscale=-1]
    % outside
    \fill[bg] (0, 0) rectangle (15, 7);
    
    % foreground shape
    \fill[fg] (1, 0) rectangle (14, 7);
    
    % first hole
    \fill[bg] (2, 1) -- ++(5, 0) -- ++(0, 2) -- ++(1, 0) -- ++(0, -2) -- ++(5, 0) -- ++(0, 5) -- ++(-1, 0) -- ++(0, -4) -- ++(-3, 0) -- ++(0, 2) -- ++(-3, 0) -- ++(0, -2) -- ++(-3, 0) -- ++(0, 4) -- ++(-1, 0) -- cycle;
    
    % second hole
    \fill[bg] (4, 3) -- ++(1, 0) -- ++(0, 2) -- ++(5, 0) -- ++(0, -2) -- ++(1, 0) -- ++(0, 3) -- ++(-7, 0) -- cycle;
    
    % connect 1st hole to outside
    \fill[bg] (13, 5) -- ++(1, 0) -- ++(0.5, 0.5) -- ++(-0.5, 0.5) -- ++(-1, 0) -- ++(-0.5, -0.5) -- cycle;
    
    % grid
    \draw[step=1](0, 0) grid (15, 7);
    
    % labels
    \draw (0.5, 0.5) node {0};
    \draw (1.5, 0.5) node {1};
    \draw (14.5, 0.5) node {0};
    \draw (2.5, 1.5) node {2};
    \draw (8.5, 1.5) node {3};
    \draw (3.5, 2.5) node {4};
    \draw (9.5, 2.5) node {5};
    \draw (4.5, 3.5) node {6};
    \draw (10.5, 3.5) node {7};
    
    % line index
    \foreach \y in {0, 1, 2, 3, 4, 5, 6} \draw[color=gray] (-1.5, 0.5+\y) node {$i = \y$};
    
    % arrows
    \SurroundedBy{1.2, 0.5}{0.8, 0.5} % 1 -> 0
    \SurroundedBy{2.2, 1.5}{1.8, 1.5} % 2 -> 1
    \SurroundedBy{8.2, 1.5}{7.8, 1.5} % 3 -> 1
    \SurroundedBy{3.2, 2.5}{2.8, 2.5} % 4 -> 2
    \SurroundedBy{9.2, 2.5}{8.8, 2.5} % 5 -> 3
    \SurroundedBy{4.2, 3.5}{3.8, 3.5} % 6 -> 4
    \SurroundedBy{10.2, 3.5}{9.8, 3.5} % 7 -> 5
    % \draw[-{Straight Barb}] (1.2, 0.5) -> (0.8, 0.5); % 1 -> 0
    % \draw[-{Straight Barb}] (2.2, 1.5) -> (1.8, 1.5); % 2 -> 1
    % \draw[-{Straight Barb}] (8.2, 1.5) -> (7.8, 1.5); % 3 -> 1
    % \draw[-{Straight Barb}] (3.2, 2.5) -> (2.8, 2.5); % 4 -> 2
    % \draw[-{Straight Barb}] (9.2, 2.5) -> (8.8, 2.5); % 5 -> 3
    % \draw[-{Straight Barb}] (4.2, 3.5) -> (3.8, 3.5); % 6 -> 4
    % \draw[-{Straight Barb}] (10.2, 3.5) -> (9.8, 3.5); % 7 -> 5
\end{scope}}\\
    
    \usebox\tikzfigure\smallskip\\
    
    \begin{tabular}{l@{ }ll@{ }l}
        \tikz[x=2em]{\SurroundedBy{1, 0}{0, 0}} & Adjacency &
        \tikz[x=2em]{\Equiv[bg]{1, 0}{0, 0}} & BG equivalence \\
        \tikz[x=2em]{\NotSurroundedBy{1, 0}{0, 0}} & Adjacency discarded &
        \tikz[x=2em]{\Equiv[fg]{1, 0}{0, 0}} & FG equivalence \\
    \end{tabular}%
    \end{tabular}%
    %\vspace{-1em}%
    \caption{Step by step example of our new B\&W labeling focusing on equivalences building and adjacency setting.}%
    \label{fig:lsl_bw}%
\end{figure}

\Cref{fig:lsl_bw} shows how our algorithm builds the equivalence table $T$ and the adjacency tree $I$ on a simple, yet complete, example.
It shows the input image with initial labels and their speculative surrounding (FG in gray and BG in white), as well as a graph representing both the equivalence table $T$ and the adjacency tree $I$.\medskip

On the first three lines ($i=0$, $i=1$ and $i=2$), five new labels are created \fglabel{1}, \bglabel{2}, \bglabel{3}, \fglabel{4} and \fglabel{5}.
Their initial adjacency is set as their speculative surrounding: $\fglabel{1} \sqsubset \bglabel{0}$, $\bglabel{2} \sqsubset \fglabel{1}$, $\bglabel{3} \sqsubset \fglabel{1}$, $\fglabel{4} \sqsubset \bglabel{2}$ and $\fglabel{5} \sqsubset \bglabel{3}$.

At $i=3$, two new labels are created with the following speculative surroundings: $\bglabel{6} \sqsubset \fglabel{4}$ and $\bglabel{7} \sqsubset \fglabel{5}$.
In addition, $\bglabel{3} \equiv \bglabel{2}$ is detected.
Consequently, the speculative surrounding of $\bglabel{3}$ is discarded in favor of $\bglabel{2} \sqsubset \fglabel{1}$.

At $i=4$, as $\fglabel{5} \equiv \fglabel{4}$, the speculative surrounding $\fglabel{5} \sqsubset \bglabel{3}$ is discarded.

At $i=5$, two new equivalences are detected: $\bglabel{2} \equiv \bglabel{0}$ and \mbox{$\bglabel{7} \equiv \bglabel{6}$}.
Consequently, the speculative surroundings of \bglabel{2} and \bglabel{7} are dropped.
The component \bglabel{0}\bglabel{2}\bglabel{3} has no more surrounding as \bglabel{0} is the exterior of the image.
While the algorithm is not capable to detect it, we can see that the surrounding $\bglabel{6} \sqsubset  \fglabel{4}$ is no more speculative and is actually \replace{correct}{final}.

At $i=6$, the last equivalence $\fglabel{4} \equiv \fglabel{1}$ is detected and the speculative surrounding $\fglabel{4} \sqsubset \bglabel{2}$ is discarded, and the surrounding $\fglabel{1} \sqsubset \bglabel{0}$ is kept.

This leads to the final state before transitive closure where all remaining surroundings ($\bglabel{6} \sqsubset \fglabel{4}$ and $\fglabel{1} \sqsubset \bglabel{0}$) are no more speculative and are actually true surroundings.
When holes are filled, the adjacency edge $\bglabel{6} \sqsubset \fglabel{4}$ is replaced by an equivalence edge $\fglabel{6} \equiv \fglabel{4}$.
Note that our algorithm actually fills hole during transitive closure and not beforehand.

\section{Benchmark \& Performance Analysis} \label{sec:benchmark_performance_analysis}
% ====================================================================================

%Dans le tableau de resultat, mettre la citation nue [] pour chaque ligne, comme Bit-Quads and loop-unrolling \cite{He2017_bitquad_unroll_euler}.

%\begin{itemize}
%\item computation of the adjacency tree (parallel algo.):\cite{DiazDelRio2020_homotopy_tree} algorithme parallel lent

%\item linear speedup, but no gain \cite{DiazDelRio2019_adjacency_tree}
%\item He Bit-Quads and loop-unrolling: \cite{He2017_bitquad_unroll_euler}
%\item closing pixel (le seul) \cite{He2015_closing_pixel}
%\item B\&W labeling en 1 passe, HCS et HCS2 \cite{He2012_HCS_hole}
%\end{itemize}

We measured the performance of our algorithms using a protocol similar to \cite{Grana2018_JRTIP}.
We tested randomly generated \nbyn{2048} images with varying density and granularity on a Skylake Gold 6126 Xeon @2.60GHz.
Grana's\cite{Grana2019_spaghetti} and Diaz'\cite{DiazDelRio2020_homotopy_tree} works have been ran and measured on our machine.
We also ran the CCA algorithms from \cite{Cabaret2014_CCL_SIPS} on our machine and kept three of them: Rosenfeld+F as the reference, He HCS2+F and LSL$_{STD}$+F as the fastest pixel-based and run-based algorithms at this time.
The other ones have been estimated from their paper.
To have comparable results across machines, we give all the results in cycles per pixel (cpp) that is the execution time multiplied by the clock frequency and divided by the number of pixels.\medskip

% -----------
\begin{table}
% -----------
    \centering
    % \begin{tabular}{l|S[table-format=2.2,table-space-text-pre={+ }]|S[table-format=1.2,table-space-text-pre={+ }]}
    %     {} & {max} & {min} \\\hline
    %     B\&W + Adjacency & 15.3 & 2.80 \\
    %     %BW1 + Adjacency & 13.3 & 2.57 \\
    %     \hline
    %     +Euler number         & {+ }0.66 & {+ }0 \\
    %     +Hole Filling         & {+ }0.74 & {+ }0 \\
    %     +Feature Computation  & {+ }10.3 & {+ }0 \\
    %     +Relabeling           & {+ }13.2 & {+ }0.59 \\
    % \end{tabular}
    \begin{tabular}{l|S[table-format=1.2,table-space-text-pre={+ }]|S[table-format=2.2,table-space-text-pre={+ }]}
        {} & {min} & {max} \\\hline
        B\&W + Adjacency & 2.80 & 15.3 \\
        %BW1 + Adjacency & 2.57 & 13.3 \\
        \hline
        +Euler number         & {+ }0    & {+ }0.66 \\
        +Hole Filling         & {+ }0    & {+ }0.74 \\
        +Feature Computation  & {+ }0    & {+ }10.3 \\
        +Relabeling           & {+ }0.59 & {+ }13.2 \\
    \end{tabular}
    \caption{Minimal and maximal processing time in \unit{cpp} of our base B\&W algorithm and extra computation for \nbyn{2048} random images.}
    \label{tab:lsl_cpp}
\end{table}
\Cref{tab:lsl_cpp} shows the minimal and maximal processing time of our new labeling algorithm.
The first line corresponds to a \emph{base} processing: foreground and background CC labeling and computing their adjacency tree.
The next lines provide the extra times to do extra computations like Euler or Hole Filling and  B\&W Feature Computation.
The extra times are the worst case we measured for doing this extra computation.
One can then estimate the total processing time for all the computations they are interested in just by adding all the extra times.

On this table, we can observe that the minimal extra time for all computations but relabeling is 0.
This is a property of run-based algorithms: those computation times depend on the number of segments -- which is 1 per line for empty images.
Euler number computation and Hole filling are really inexpensive using our approach.
On the opposite, relabeling is very expensive, almost doubling the total time, and should be avoided if not required.\medskip

% ------------
\begin{figure}
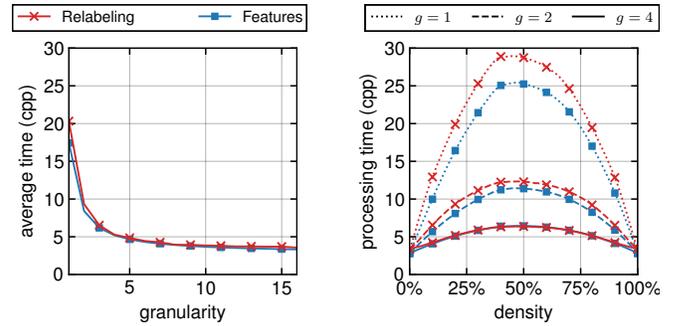

% ------------
    \includepgf[scale=0.70]{plots}{legend-bw}\hfill
    \includepgf[scale=0.70]{plots}{legend-granularity}\smallskip
    
    \subcaptionbox{Average processing time over foreground density for $g \in \interval{1}{16}$}{\includepgf[scale=0.70]{plots}{skx-average-v0}\hspace{1ex}}
    \hfill
    \subcaptionbox{Processing time depending on foreground density for $g \in \{1, 2, 4\}$}{\hspace{1ex}\includepgf[scale=0.70]{plots}{skx-density-v0}}%
    \caption{Processing time in \unit{cpp} of hole filling with either feature computation or relabeling.}%
    \label{plot:bw}%
\end{figure}
\Cref{plot:bw} shows the processing time of the labeling with hole filling and either relabeling or feature computation depending on both granularity and density.
We can see that the processing time quickly decreases with higher granularity.
Random images with $g = 1$ are highly artificial but are useful to stress the algorithms where they appear to be slowest.
Our labeling is much faster at $g = 4$ with an average time of $5.3 \unit{cpp}$.
%This is interesting because natural images seem to be processed roughly as fast as those $g = 4$ random images.\todo{reformulate this last sentence?}\medskip
%
This is interesting because natural images (like SIDBA database) usually have an average  processing time close to the processing time of random images with granularity greater than 4, according to \cite{Cabaret2014_CCL_SIPS}.\medskip % average plutot sur random images? -> oui
% en verifiant dans SIPS, c'est meme g=8, et pour STD et pour RLE
% comme cela on est bon

% -----------
\begin{table}
% -----------
\newcommand{\alignright}{\hspace{0pt plus 1filll}}%
    % \begin{tabularx}{\linewidth}{@{}L@{}c|S@{\hspace{1ex}}S|c@{\hspace{1em}}c@{}}
    %     \multirow{2}{*}{Algorithm} & \multirow{2}{*}{Compute} & \multicolumn{2}{c|}{Their} & \multicolumn{2}{c}{Our} \\
    %     {} & {} & {max} & {min} & {max} & {min} \\\hline\hline
    %     {Grana\cite{Grana2019_spaghetti}} & WR \alignright\textcolor{gray}{+BA} & 25.7 & 3.40 & \multirow{2}{*}{28.9} & \multirow{2}{*}{3.41} \\
    %     Diaz\cite{DiazDelRio2020_homotopy_tree} & BWAR \alignright\null & 59.0 & 18.4 & {} & {} \\\hline
        
    %     He BW\cite{He2013_HCS_hole} & BWER \alignright\textcolor{gray}{+A} & 79.7 & 9.00 & \multirow{2}{*}{29.0} & \multirow{2}{*}{3.41} \\
    %     He combined\cite{He2015_closing_pixel} & WER \alignright\textcolor{gray}{+BA} & 48.0 & 16.6 & {} & {} \\\hline
        
    %     He run-based\cite{He2015_euler_run} & E \alignright\textcolor{gray}{+BWA} & 36.5 & 5.54 & \multirow{2}{*}{15.9} & \multirow{2}{*}{2.80} \\
    %     He bit-quad\cite{He2017_bitquad_unroll_euler} & E \alignright\textcolor{gray}{+BWA} & 23.7 & 2.87 & {} & {} \\\hline
        
    %     Rosenfeld+F$^*$ \cite{Cabaret2014_CCL_SIPS} & WF \alignright\textcolor{gray}{+BA} & 48.0 & 4.68 & \multirow{3}{*}{24.8} & \multirow{3}{*}{2.80} \\
    %     He HCS2+F$^*$ \cite{Cabaret2014_CCL_SIPS} & WF \alignright\textcolor{gray}{+BA} & 38.6 & 4.33 & {} & {} \\
    %     Classical LSL$_{STD}$+F\cite{Cabaret2014_CCL_SIPS} & WF \alignright\textcolor{gray}{+BA} & 14.4 & 2.69 & {} & {} \\
        
    % \end{tabularx}\vspace{1ex}
    \begin{tabularx}{\linewidth}{@{}L@{}|>{\footnotesize}c@{\hspace{1ex}}S[table-format=2.2]@{\hspace{1em}}S[table-format=2.1]|>{\footnotesize}c@{\hspace{1ex}}c@{\hspace{0.9em}}c@{}}
        \multirow{2}{*}{Algorithm} & \multicolumn{3}{c|}{Their} & \multicolumn{3}{c}{Our} \\
        {} & {compute} & {min} & {max} & {compute} & {min} & {max} \\\hline\hline
        {Grana\cite{Grana2019_spaghetti}} & WR & 3.40 & 25.7 & \multirow{2}{*}{BWAR} & \multirow{2}{*}{3.41} & \multirow{2}{*}{28.9} \\
        Diaz\cite{DiazDelRio2020_homotopy_tree} & BWAR & 18.4 & 59.0 & {} & {} & {} \\\hline
        
        He BW\cite{He2013_HCS_hole} & BWER & 9.00 & 79.7 & \multirow{2}{*}{BWAER} & \multirow{2}{*}{3.41} & \multirow{2}{*}{29.0} \\
        He combined\cite{He2015_closing_pixel} & WER & 16.6 & 48.0 & {} & {} & {} \\\hline
        
        He run-based\cite{He2015_euler_run} & E & 5.54 & 36.5 & \multirow{2}{*}{BWAE} & \multirow{2}{*}{2.80} & \multirow{2}{*}{15.9} \\
        He bit-quad\cite{He2017_bitquad_unroll_euler} & E & 2.87 & 23.7 & & {} & {} \\\hline
        
        Rosenfeld+F$^*$ \cite{Cabaret2014_CCL_SIPS} & WF & 4.68 & 48.0 & \multirow{3}{*}{BWAF} & \multirow{3}{*}{2.80} & \multirow{3}{*}{24.8} \\
        He HCS2+F$^*$ \cite{Cabaret2014_CCL_SIPS} & WF & 4.33 & 38.6 & {} & {} \\
        LSL$_{STD}$+F\cite{Cabaret2014_CCL_SIPS} & WF & 2.69 & 14.4 & {} & {} \\
        
    \end{tabularx}\vspace{1ex}
    \begin{scriptsize}
        \begin{tabularx}{\linewidth}{@{}c@{: }l@{ }l@{ }l@{}X@{}c@{: }l@{}X@{}c@{: }l@{}}
            B & Black & labeling & (BG) && A & Adjacency tree && F & Feature Computation \\
            W & White & labeling & (FG) && E & Euler number   && R & Relabel\\
            $*$ & \multicolumn{9}{@{}l@{}}{standard CCL algorithms transformed into 1-pass CCA algorithms (features only)}
        \end{tabularx}%
    \end{scriptsize}%
    \caption{Performance comparison between State-of-the-Art algorithms (``Their'') and this work (``Our''). The ``compute'' columns show what is computed by the algorithms. ``min'' and ``max'' columns show the minimum and maximum processing time in \unit{cpp} measured for each algorithm.}
    \label{tab:state_of_the_art}
\end{table}
In \cref{tab:state_of_the_art}, each State-of-the-Art algorithm are compared to one configuration of our new algorithm that computes at least as much.
The execution time of our algorithm is close to the best classical foreground CCL algorithm \cite{Grana2019_spaghetti} despite computing background labels and adjacency tree in addition.
This is possible thanks to our light-weight adjacency-tree computation and the use of segments that allows a \emph{symmetric} computation of BG and FG components.

Our work outperforms both existing B\&W algorithms \cite{DiazDelRio2020_homotopy_tree,He2013_HCS_hole} or CCL algorithm with Euler number computation \cite{He2015_closing_pixel}.
While we compute much more, Euler computation is faster than dedicated algorithms \cite{He2015_euler_run, He2017_bitquad_unroll_euler}.
Our run-based algorithm is faster than pixel-based CCA algorithms, like already observed in \cite{Cabaret2014_CCL_SIPS} with classical LSL.
Classical LSL$_{STD}$+F -- the base of our algorithm -- remains obviously faster.
The most noticeable difference is for the maximum processing time: our BW algorithm needs to process twice as much segments than classical LSL.

%And as previously obserbed by Cabaret \emph{et al} \cite{Cabaret2014_CCL_SIPS}, a run-based algorithm like LSL still outperforms pixel-based CCA algorithms. \medskip

% This benchmark shows that our new LSL algorithm -- that is able to compute as much as all existing algorithms (B\&W labeling, Euler, adjacency tree) -- is faster than all of them and close to fastest classic black labeling \cite{Grana2019_spaghetti} but doing a double labeling.

%\textcolor{darkgreen}{bla bla bla bla bla bla bla bla bla bla bla bla bla bla bla bla bla bla bla bla bla bla bla bla bla bla bla bla bla bla bla bla bla bla bla bla bla bla bla bla bla bla bla bla bla bla bla bla bla bla bla bla bla bla bla bla bla bla bla bla bla bla bla bla bla bla bla bla bla bla bla bla bla bla bla bla bla bla bla bla bla bla bla bla bla bla bla bla bla bla bla bla bla bla bla bla bla bla bla bla bla bla bla bla bla bla bla bla bla bla bla bla bla bla bla bla bla bla bla bla bla bla bla bla bla bla bla bla bla bla bla bla bla bla}
% =========================================
\section{Conclusion} \label{sec:conclusion}
% =========================================

In this article, we have introduced a new connected component labeling and analysis algorithm that is able to do in one single pass of the image, both the Euler number computation but also a double foreground and background labeling with the adjacency tree computation. The modified transitive closure algorithm enables an efficient hole processing: holes can be filled and the surrounding connected components are updated on-the-fly.
\new{Our approach can easily be adapted to other connected operators like filtering out components based on their statistical features.}\medskip

As far as we know our algorithm is faster than B\&W labeling algorithms and algorithms computing features related to hole processing. 
%all published algorithms related to black and white topology.

%\footnotesize
\bibliographystyle{IEEEbib}
\bibliography{ref}
%\balance

\end{document}